\pdfoutput=1

\documentclass[11pt]{article}

\usepackage[]{acl}
\usepackage{multirow}

\usepackage{times}
\usepackage{latexsym}

\usepackage{graphicx}
\usepackage{amsmath}
\usepackage{hyperref}

\usepackage[T1]{fontenc}

\usepackage[utf8]{inputenc}

\usepackage{microtype}

\title{Mitigating Gender Stereotypes in Hindi and Marathi}

\newcommand\blfootnote[1]{%
  \begingroup
  \renewcommand\thefootnote{}\footnote{#1}%
  \addtocounter{footnote}{-1}%
  \endgroup
}

\author{Neeraja Kirtane\thanks{~~First author}\\
  Manipal Institute of Technology \\
  MAHE, Manipal, India \\
  \texttt{kirtane.neeraja@gmail.com} \\\And
  Tanvi Anand \\
  University of Texas, Austin \\
   Texas, USA \\
  \texttt{tanvianand@gmail.com} \\}

\begin{document}
\maketitle
\begin{abstract}
As the use of natural language processing increases in our day-to-day life, the need to address gender bias inherent in these systems also amplifies. This is because the inherent bias interferes with the semantic structure of the output of these systems while performing tasks like machine translation. While research is being done in English to quantify and mitigate bias, debiasing methods in Indic Languages are either relatively nascent or absent for some Indic languages altogether. Most Indic languages are gendered, i.e., each noun is assigned a gender according to each language's grammar rules. As a consequence, evaluation differs from what is done in English. This paper evaluates the gender stereotypes in Hindi and Marathi languages. The methodologies will differ from the ones in the English language because there are masculine and feminine counterparts in the case of some words. We create a dataset of neutral and gendered occupation words, emotion words and measure bias with the help of Embedding Coherence Test (ECT) and Relative Norm Distance (RND). We also attempt to mitigate this bias from the embeddings. Experiments show that our proposed debiasing techniques reduce gender bias in these languages.

\end{abstract}

\blfootnote{\textit{Accepted at the GeBNLP workshop at NAACL 2022}}

\section{Introduction}
Word embeddings are used in most natural language processing tools. Apart from capturing semantic information, word embeddings are also known to capture bias in society \citep{bolukbasi2016man}. While most research has been focused on languages like English, less research has been done on low-resource languages and languages that have a grammatical gender. A language with grammatical has a gender associated with every noun irrespective of whether the noun is animate or inanimate, e.g., a river in Hindi has feminine gender. In contrast, words like writer have masculine and feminine counterparts. This gender association affects the pronouns, adjectives, and verb forms used during sentence construction. Grammatical genders in Hindi are masculine and feminine. In Marathi, there additionally exists a third neutral gender as well. Spoken by more than 600 million people, Hindi is the 3rd most spoken language in the world. Marathi is the 14th most spoken language with approximately 120 million speakers \footnote{\url{https://www.mentalfloss.com/article/647427/most-spoken-languages-world}}. Given the expanse and the amount of people speaking these languages, it is essential to address the bias introduced by the computational applications of these languages. \par

We create a dataset of gendered and neutral occupation titles. We also create a dataset with words of different emotions like anger, fear, joy, and sadness. First, we identify existing gender bias by defining a subspace that captures the gender information. There are several ways to find this information, such as Principal Component Analysis (PCA) and Relational Inner Product Association (RIPA). We use the existing metrics for evaluation: Embedding Coherence Test \citep{dev2019attenuating}, Relative Norm Distance \citep{rnd}. We modify these formulas so that they are correctly applicable to these gendered languages. We perform our experiments on the FastText word embeddings.\par
Next, we mitigate the gender bias found by the aforementioned using two approaches: Projection and Partial Projection. In summary, the key contributions of this paper are:
\begin{enumerate}
    \item Dataset of emotions, and gendered and neutral occupations in Hindi and Marathi.
    \item Methods to quantify the bias present in
Hindi and Marathi word embeddings using the above dataset.
    \item Mitigate the bias through existing debiasing techniques.
\end{enumerate}

\section{Related work}
Previous work to quantify and measure bias was done by \citet{bolukbasi2016man}. They tried to find out a gender subspace by using gender-definition pairs. They proposed a hard de-biasing method that identifies the gender subspace and tries to remove its components from the embeddings.\par
The maximum amount of research on gender bias is being done in English, which is not gendered \citep{stanczak2021survey}. Languages like Spanish or Hindi have a grammatical gender, i.e., every noun is assigned a gender. \citet{zhou} was one of the first papers to examine bias in languages with grammatical gender like  French and Spanish. They used a modified version of the Word Embedding Association Test (WEAT) \citep{WEAT} to quantify the bias.\par
\citet{sun-etal-2019-mitigating} suggested mitigation techniques to remove gender bias like data augmentation, gender-swapping, and hard de-biasing according to the downstream task in NLProc. \par
Being low-resource languages, there is less research done in languages like Hindi and Marathi. 
Previous work in Indic Languages was done by \citet{inproceedings} where they built an SVM classifier to identify the bias and classify it. The problem with this method is that it needs a labeled gender dataset beforehand to train the classifier. Recent work by \citet{ramesh2021evaluating} tries to find out bias in English-Hindi machine translation. They implement a modified version of the TGBI metric based on grammatical considerations for Hindi. TGBI metric is used to detect and evaluate gender bias in Machine Translation systems. \citet{malik2021socially} measure Hindi specific societal biases like religion bias and caste bias along with gender bias.

\section{Data}
In \citet{bolukbasi2016man}, the authors have compiled a list of professions in English and tried to find bias in them. Similarly, we compile a list of 166 professions, each in Hindi and Marathi languages. 
We split the professions into two parts, first is gender-neutral $P_{neu}$ and the other is gendered $P_{gen}$. 
Similarly, we create a list of words of different emotions similar to the one in \citet{saif} in Hindi and Marathi languages. The emotions are broadly classified into four types: anger, fear, joy, and sadness. \par
\begin{figure}[htp]
    \centering
    \includegraphics[width=7cm]{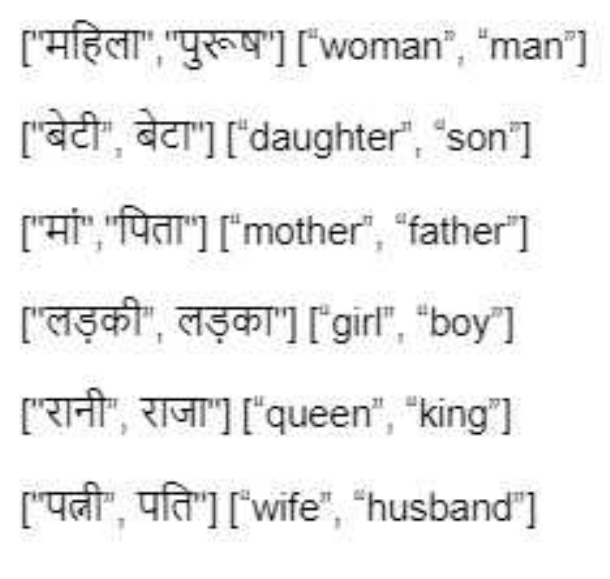}
    \caption{Example of seed words in Hindi and their English translation}
    \label{fig:nb}
\end{figure}

 We have verified this data with the help of 5 independent native speakers of these languages. 
We also create pair of feminine and masculine seed word pairs in both the languages to identify the gender subspace. For example: queen, king. We call them target words $T$. Target words for Hindi Language is shown in figure \ref{fig:nb}. 
The dataset is available here \footnote{\url{https://github.com/neeraja1504/GenderBias_corpus}} \par
We test the bias on our data using FastText embeddings. FastText is a word embedding method that extends the word2vec model. Instead of learning vectors for words directly, FastText represents each word as an n-gram of characters. This helps capture the meaning of shorter words and allows the embeddings to understand suffixes and prefixes. A skip-gram model is trained to learn the embeddings once the word has been represented using character n-grams \cite{DBLP:journals/corr/BojanowskiGJM16}.\par 
Morphology is the field of linguistics that studies the internal structure of words. Morphologically rich languages refer to languages that contain a substantial amount of grammatical information \citep{comrie1999grammatical}. Indic languages are morphologically rich because of the existence of a large number of different word forms. FastText embeddings are the best choice for Indian Languages as they are capable of capturing and integrating sub-word information using character n-gram embeddings during training \citep{kunchukuttan2020ai4bharat}.

\begin{table*}
\centering
\begin{tabular}{|l|l|l|ll|ll|}
\hline
Emotion                  & Metrics(\%) & Baseline & \multicolumn{2}{l|}{Projection}      & \multicolumn{2}{l|}{Partial Projection} \\ \hline
                         &         &          & \multicolumn{1}{l|}{PCA}    & RIPA   & \multicolumn{1}{l|}{PCA}      & RIPA    \\ \hline
\multirow{2}{*}{Anger}   & ECT     & 57.2   & \multicolumn{1}{l|}{94.5} & 38.1  & \multicolumn{1}{l|}{\textbf{99.1}}        & 99.0   \\ \cline{2-7} 
                         & RND     & 2.0    & \multicolumn{1}{l|}{1.1}  & 0.4  & \multicolumn{1}{l|}{0.7}    & \textbf{0.3}   \\ \hline
\multirow{2}{*}{Fear}    & ECT     & 86.8   & \multicolumn{1}{l|}{97.8} & 77.4  & \multicolumn{1}{l|}{\textbf{99.4}}    & 98.9   \\ \cline{2-7} 
                         & RND     & 2.6    & \multicolumn{1}{l|}{1.2}  & 0.3  & \multicolumn{1}{l|}{0.8}    & \textbf{0.2}   \\ \hline
\multirow{2}{*}{Joy}     & ECT     & 75.2    & \multicolumn{1}{l|}{96.1} & 81.1  & \multicolumn{1}{l|}{\textbf{99.5}}    & 99.2   \\ \cline{2-7} 
                         & RND     & 2.5    & \multicolumn{1}{l|}{1.2}  & 0.5  & \multicolumn{1}{l|}{0.8}    & \textbf{0.3}   \\ \hline
\multirow{2}{*}{Sadness} & ECT     & 63.1    & \multicolumn{1}{l|}{88.4}  & 56.1 & \multicolumn{1}{l|}{\textbf{99.4}}    & 98.3   \\ \cline{2-7} 
                         & RND     & 3.8    & \multicolumn{1}{l|}{1.7}  & 0.7  & \multicolumn{1}{l|}{0.9}    & \textbf{0.4}   \\ \hline
\end{tabular}
\caption{\label{table3}
Hindi Emotion Results (Principal Component Analysis (PCA), Relational Inner Product Association (RIPA))
}
\end{table*}

\begin{table*}
\centering
\begin{tabular}{|l|l|l|ll|ll|}
\hline
Emotion                  & Metrics(\%) & Baseline & \multicolumn{2}{l|}{Projection}      & \multicolumn{2}{l|}{Partial Projection} \\ \hline
                         &         &          & \multicolumn{1}{l|}{PCA}    & RIPA   & \multicolumn{1}{l|}{PCA}      & RIPA    \\ \hline
\multirow{2}{*}{Anger}   & ECT     & 37.9   & \multicolumn{1}{l|}{58.2}  & 52.7  & \multicolumn{1}{l|}{\textbf{96.1}}    & 93.9   \\ \cline{2-7} 
                         & RND     & 0.61    & \multicolumn{1}{l|}{0.53} & 0.10  & \multicolumn{1}{l|}{0.33}   & \textbf{0.09}  \\ \hline
\multirow{2}{*}{Fear}    & ECT     & 72.6   & \multicolumn{1}{l|}{74.0}  & 71.2  & \multicolumn{1}{l|}{\textbf{96.4}}    & 93.6   \\ \cline{2-7} 
                         & RND     & 0.50    & \multicolumn{1}{l|}{0.41} & 0.11  & \multicolumn{1}{l|}{0.29}   & \textbf{0.08}  \\ \hline
\multirow{2}{*}{Joy}     & ECT     & 57.3   & \multicolumn{1}{l|}{76.2}  & 58.2  & \multicolumn{1}{l|}{\textbf{93.4}}    & 92.5   \\ \cline{2-7} 
                         & RND     & 0.60    & \multicolumn{1}{l|}{0.61} & 0.13 & \multicolumn{1}{l|}{0.39}   & \textbf{0.11}  \\ \hline
\multirow{2}{*}{Sadness} & ECT     & 69.6    & \multicolumn{1}{l|}{80.2}  & 68.9  & \multicolumn{1}{l|}{\textbf{99.1}}    & 96.9   \\ \cline{2-7} 
                         & RND     & 0.42   & \multicolumn{1}{l|}{0.37} & 0.08 & \multicolumn{1}{l|}{0.25}   & \textbf{0.07}  \\ \hline
\end{tabular}
\caption{\label{table4} Marathi Emotion Results}
\end{table*}

\section{Methodology}
\subsection{Bias Statement}
Various definitions of bias exist and vary in research as explained in the paper \citep{blodgett2020language}. Our work focuses on stereotypical associations between masculine and feminine gender and professional occupations and emotions in FastText word embeddings. The classic example of "He is a doctor" and "She is a nurse" comes into play here. It is especially harmful to the representation of minority communities, since these stereotypes often end up undermining these communities \citep{moss2012science}. Downstream NLP applications learn from these stereotypes, and the risk of discrimination on the basis of gender in this case keeps seeping further into the system.


Our work tries to de-correlate gender with occupation and emotions, which will help reduce bias in these systems.
\subsection{Quantifying bias for Occupations and Emotions}
We use the following methods to quantify the bias before and after debiasing. $M_{gen}$ is used for gendered attributes like gendered occupations. $M_{neu}$ is used for neutral attributes like emotions and neutral occupations. We use these two different methods because our data has two different parts — gendered and neutral. 
\subsubsection{For neutral occupations: $M_{neu}$}
\begin{enumerate}
    \item \textbf{ECT-n}: \citet{dev2019attenuating} use this test to measure bias. We use the target word pairs $T$, and the neutral attributes list $P_{neu}$. We separate the target word pairs into masculine and feminine-targeted words respectively. For each of the pairs ${\vec{mi},\vec{fi}}$ in $T$ we create two means $\vec{a1}$ and $\vec{a2}$.
    \begin{equation}
        \vec{a1}= \frac{ 1 }{|M|} \sum_{\vec{m} \in M} \vec{m}
        \label{1}
    \end{equation}
     \begin{equation}
        \vec{a2}= \frac{ 1 }{|F|} \sum_{\vec{f} \in F} \vec{f}
        \label{2}
    \end{equation}
    $M$ are masculine word embeddings, $F$ are feminine word embeddings of the target word pairs $T$.
     We then create two arrays, one with the cosine similarity between the neutral word embeddings and $\vec{a1}$, the other with the neutral word embeddings and $\vec{a2}$. We calculate the Spearman correlation between the rank orders of these two arrays found. Spearman rank correlation is a non-parametric test that is used to measure the degree of association between two variables. Higher the correlation, the less the bias. The range of the correlation is \([-1,1]\). Ideally, the correlation should be equal to one as the professions or emotions should not depend upon gender. Debiasing should bring the value closer to one.
    \item \textbf{RND-n}: Relative Norm Distance was first used by \citet{rnd}. It captures the relative strength of the association of a neutral word with respect to two groups.
    As shown in equation \ref{equation:3} we average the masculine and feminine-targeted words in $M$, $F$ in $T$ respectively. For every attribute, $\vec{p}$ in $P_{neu}$, emotions we find the norm of the average of the target words and the attribute $\vec{p}$. The higher the value of the relative norm, the more biased our professions and emotions are. Debiasing should reduce this value and bring it closer to zero. 
    \begin{equation}
        \sum_{\vec{p} \in P_{neu}} (||avg(M)-\vec{p}||_2 - ||avg(F)-\vec{p}||_2)
        \label{equation:3}
    \end{equation}
\end{enumerate}

\subsubsection{For gendered occupations:$M_{gen}$}
\begin{enumerate}
    \item \textbf{ECT-g}: We use the target word pairs $T$ and the gendered professions list $P_{gen}$. Using $\vec{a1}$ found in equation \ref{1} and $\vec{a2}$ found in equation \ref{2}. $P_{gen}$ has masculine and feminine profession word pairs. We create two arrays, one with cosine similarity of masculine profession word embeddings and $\vec{a1}$. The other with the cosine similarity of feminine profession word embeddings and $\vec{a2}$. We calculate the Spearman correlation of the rank of these two arrays.\par
    Ideally, there should be a high correlation between these arrays. The masculine profession words' cosine similarity with masculine target words should equal feminine profession words’ cosine similarity with feminine target words.
    The range of the correlation is \([-1,1]\). Higher the correlation, the less the bias. Debiasing should bring the value closer to one.
    \item \textbf{RND-g}:  As shown in equation \ref{4} we average the masculine and feminine-targeted words in $M$, $F$ in $T$, respectively. For every attribute pair $\vec{p1}$ and $\vec{p2}$ in $P_{gen}$ we find the norm of the average of the masculine target words and $\vec{p1}$ , feminine target words and $\vec{p2}$. The higher the value of the relative norm, the more biased the professions are. Debiasing should reduce this value and bring it closer to zero.
    \begin{equation}
        \sum_{\vec{p1},\vec{p2} \in P} (||avg(M)-\vec{p1}||_2 - ||avg(F)-\vec{p2}||_2)
        \label{4}
    \end{equation}
\end{enumerate}

\subsection{Debiasing techniques}
\subsubsection{Finding out the gender subspace}
We need a vector $\vec{v_{b}}$ that represents the gender direction. We find this in the following ways: using RIPA and PCA.
\begin{enumerate}
    \item \textbf{RIPA}:  \citet{ripa} first used this subspace to capture gender information. We define a bias vector $\vec{v_{b}}$ which defines the gender direction. Given the target set $T$ containing masculine and feminine words, for each $T_{j}$ in $T$, we find out $T_{f}-T_{m}$ and stack them to create an array. $T_{f}$ is the feminine word embedding, $T_{m}$ is the masculine word embedding. We find the first principal component using Principal Component Analysis (PCA) of the array found above. This component captures the gender information of the given embeddings.
    \item \textbf{PCA}: In this method, given $T$, we find out the average $a$ of the masculine and feminine word embeddings for each given pair. We then compute $T_{f} - a$ and $T_{m} - a$ for each $T_{j}$ in $T$. We stack them into an array and find out the first component using the PCA of the above array. 
    
\end{enumerate}

\subsubsection{Debiasing methods}
  
 

\begin{table}[]
\begin{tabular}{|l|l|ll|ll|}
\hline
Metric & Base  & \multicolumn{2}{l|}{Projection}    & \multicolumn{2}{l|}{Partial Proj.}  \\ \hline
   (\%)     &       & \multicolumn{1}{l|}{PCA}   & RIPA  & \multicolumn{1}{l|}{PCA}   & RIPA  \\ \hline
ECT-n   & 86.0 & \multicolumn{1}{l|}{95.6} & 81.4 & \multicolumn{1}{l|}{\textbf{99.7}} & 98.9 \\ \hline
RND-n   & 40.4 & \multicolumn{1}{l|}{19.3} & 6.2 & \multicolumn{1}{l|}{10.9} & \textbf{6.2} \\ \hline
ECT-g   & 69 & \multicolumn{1}{l|}{71.4} & 85.7 & \multicolumn{1}{l|}{\textbf{90.4}} & 88.0 \\ \hline
RND-g  & 1.79 & \multicolumn{1}{l|}{1.81}  & 1.79  & \multicolumn{1}{l|}{\textbf{1.52}}  & 1.70  \\ \hline
\end{tabular}
\caption{\label{table1} Results for Hindi occupations. RND-g is not in \% for better readability} 
\end{table}

\begin{table}[]
\begin{tabular}{|l|l|ll|ll|}
\hline
Metric & Base  & \multicolumn{2}{l|}{Projection}    & \multicolumn{2}{l|}{Partial Proj}  \\ \hline
   (\%)     &       & \multicolumn{1}{l|}{PCA}   & RIPA  & \multicolumn{1}{l|}{PCA}   & RIPA  \\ \hline
ECT-n   & 51.4 & \multicolumn{1}{l|}{60.6} & 53.6 & \multicolumn{1}{l|}{\textbf{99.7}} & 96.5 \\ \hline
RND-n   & 3.1 & \multicolumn{1}{l|}{3.0} & 3.0 & \multicolumn{1}{l|}{01.8} & \textbf{0.5} \\ \hline
ECT-g   & 42.5 & \multicolumn{1}{l|}{21.5} & 23.8 & \multicolumn{1}{l|}{64.2} & \textbf{76.2} \\ \hline
RND-g  & 2.58  & \multicolumn{1}{l|}{2.50}  & 2.54  & \multicolumn{1}{l|}{2.02}  & \textbf{1.97}  \\ \hline
\end{tabular}
\caption{\label{table2} Results for Marathi occupations. RND-g is not in \% for better readability} 
\end{table}

\citet{bolukbasi2016man} used Hard Debiasing to mitigate bias. However, in \citet{gonen2019lipstick}, they show how this method is ineffective in debiasing the embeddings. Here we use more straightforward methods to debias our data. 
\begin{enumerate}
    \item \textbf{Projection}: One way to remove bias is to make all the vectors orthogonal to the gender direction. Therefore, we remove the component of $\vec{v_{b}}$ from all the vectors. This ensures that there is no component along the gender direction.  
    \begin{equation}
        \vec{w}=\vec{w}-(\vec{w} \cdot \vec{v_{b}})\vec{v_{b}}
    \end{equation}
  
    \item \textbf{Partial Projection}: One problem with the debiasing using linear projection is that it changes some word vectors which are gendered by definition, e.g., king, queen. Let \(\mu  = \frac{1 }{ m} \sum_{j=1}^{m} \mu j\) where  \(\mu j = \frac{1}{|T_{j}|} \sum_{t \in T_{j}} t\) be the mean of a target pair. Here m is the length of $T$. We suggest the new vector as shown in equation \ref{7}. This is similar to the linear projection approach, but instead of zero magnitude along the gender direction, we project a magnitude of constant \(\mu\) along with it. This adds a constant to the debiasing term.
    \begin{equation}
        \vec{w}=\vec{w}-(\vec{w} \cdot \vec{v_{b}})\vec{v_{b}} + \mu
        \label{7}
    \end{equation}
\end{enumerate}

\section{Results and Discussion}

Table \ref{table3} and \ref{table4} show results for the emotions in Hindi and Marathi respectively. We observe that anger is the most biased in both languages according to the ECT metric as it has the lowest value. Amongst the debiasing techniques, we see that partial projection with RIPA works the best for the ECT metric and partial projection with PCA works the best for the RND metric.\par
 ECT-n and RND-n are results for neutral occupations, and ECT-g and RND-g are for gendered occupations. Table \ref{table1} shows the results for both gendered and neutral occupations in Hindi. We see that partial projection We see that for neutral occupations, partial projection with PCA works the best for ECT and partial projection with RIPA works the best for RND. For gendered occupations, we see that partial projection with PCA works the best for both ECT and RND. \par
 Table \ref{table2} shows the results for both gendered and neutral occupations in Marathi. We see that the best results are obtained for neutral occupations with partial projection with PCA for ECT and partial projection with RIPA for RND. For gendered occupations, we see that we get the best results with partial projection with RIPA for ECT and RND.\par
 
 However, we observe some anomalies in the results when projection debiasing method is used. We hypothesize that completely removing the gender information changes some vectors, which are masculine or feminine, by the grammatical definition of the gender. For example, words like king, grandfather and boy which are masculine by the grammatical definition of gender should preserve their gender information. Hence we note that partial projection performs the best because it has a gender component to it.

\section{Conclusion and Future work}
In this paper, we attempted to find gender stereotypes on occupations and emotions and tried to debias them. Embedding Coherence Test and Relative Norm Distance were used as a bias metric in the gender subspace. The debiasing methods used were projection and partial projection. But we see that partial projection as a debiasing method works the best in most cases.\par 
Future work could include trying out these techniques on downstream tasks and checking the performance before and after debiasing. The main problem with experimenting on downstream tasks is the availability of datasets in these languages. We would also like to experiment with debiasing contextual embeddings and large language models. Apart from that we would also like to address other types of bias like religion, social and cultural, which are particularly inherent in Hindi and Marathi.

\bibliography{anthology,custom}
\bibliographystyle{acl_natbib}



\end{document}